\documentclass[11pt]{article}

\usepackage[final]{acl}

\usepackage{times}
\usepackage{latexsym}

\usepackage[T1]{fontenc}
\usepackage[utf8]{inputenc}

\usepackage{microtype}

\usepackage{inconsolata}

\usepackage{graphicx}

\usepackage{amsmath,graphicx,hyperref}
\usepackage{cite}
\usepackage{amssymb,amsfonts}
\usepackage{algorithmic}
\usepackage{textcomp}
\usepackage{xcolor}
\usepackage{tcolorbox}
\usepackage{booktabs}
\usepackage{makecell}
\usepackage{multirow}
\usepackage{xcolor,paralist}
\usepackage[flushleft]{threeparttable}
\title{Rethinking Role-Playing Evaluation: \\Anonymous Benchmarking and A Systematic Study of Personality Effects} % When Identity is Unknown: Self-Generated Personality Enables Stronger Anonymous Role-Playing

\author{Ji-Lun Peng$^{\star \dagger}$ \quad Yun-Nung Chen$^{\star}$\\
        $^{\star}$National Taiwan University, Taipei, Taiwan\\
     $^{\dagger}$Academia Sinica, Taipei, Taiwan\\
     \texttt{r13946006@ntu.edu.tw}\quad\texttt{y.v.chen@ieee.org}}

\begin{document}
\maketitle
\begin{abstract}
Large Language Models (LLMs) have shown remarkable potential in developing role-playing agents (RPAs). However, current evaluation frameworks rely heavily on well-known fictional characters, raising a critical concern: models may be leveraging their internal training memory of these characters rather than demonstrating role-playing capabilities. 
This reliance often leads to significant performance degradation when RPAs encounter unseen or out-of-distribution personas.
To address this, we propose a more rigorous evaluation protocol designed to decouple role-playing proficiency from character recognition. 
Our experiments across multiple benchmarks demonstrate that \emph{anonymizing} characters degrades performance, confirming that name exposure provides implicit cues that mask a model's true capability.
To mitigate this, we investigate diverse personality augmentation as a method to enhance role fidelity in anonymous settings. We systematically analyze the impact of various personality-description methods on agent behavior and consistency.
Our results show that incorporating personality information consistently improves RPA performance. This work establishes a more equitable evaluation standard and validates a scalable, personality-enhanced framework for constructing robust RPAs.\footnote{Code: \url{https://github.com/MiuLab/RolePlayBench}}
\end{abstract}

\section{Introduction}

LLMs have demonstrated strong performance across diverse tasks~\citep{peng2024survey}.
A recent and significant application involves their use as RPAs, where they simulate specific professions~\citep{wang2023unleashing} and characters~\citep{shao2023characterllmtrainableagentroleplaying}. This approach enhances the human-likeness of interactions, with character simulation being a major application area~\citep{tseng2024two,chen2024personapersonalizationsurveyroleplaying, he2025crab, wu2025raiden}.

Current research on role-playing agents primarily focuses on simulating fictional characters, such as those from novels and films, prompting language models to generate text consistent with a predefined character style \citep{yu2025beyond, qin2025r, zhang2025revealing, wang2025characterbox}. This line of work includes the construction of various benchmark datasets for evaluating role-playing capabilities \citep{wang2025characterbox, dai2024mmrole, wu2025raiden}, as well as methodological advancements designed to enhance LLMs' role-playing performance \citep{gao2025tailorrpa, yang2025hycora, ye2025cpo, liu2025cogdual}. Enabling LLMs to simulate fictional characters has practical applications in avatar-based interactions\footnote{\url{https://character.ai}}
 and interactive gaming environments \citep{zhang2025omnicharacter}, thereby extending virtual characters beyond their original narrative contexts.
 
\begin{figure*}[hbt!]
\centering
\includegraphics[width=0.95\textwidth]{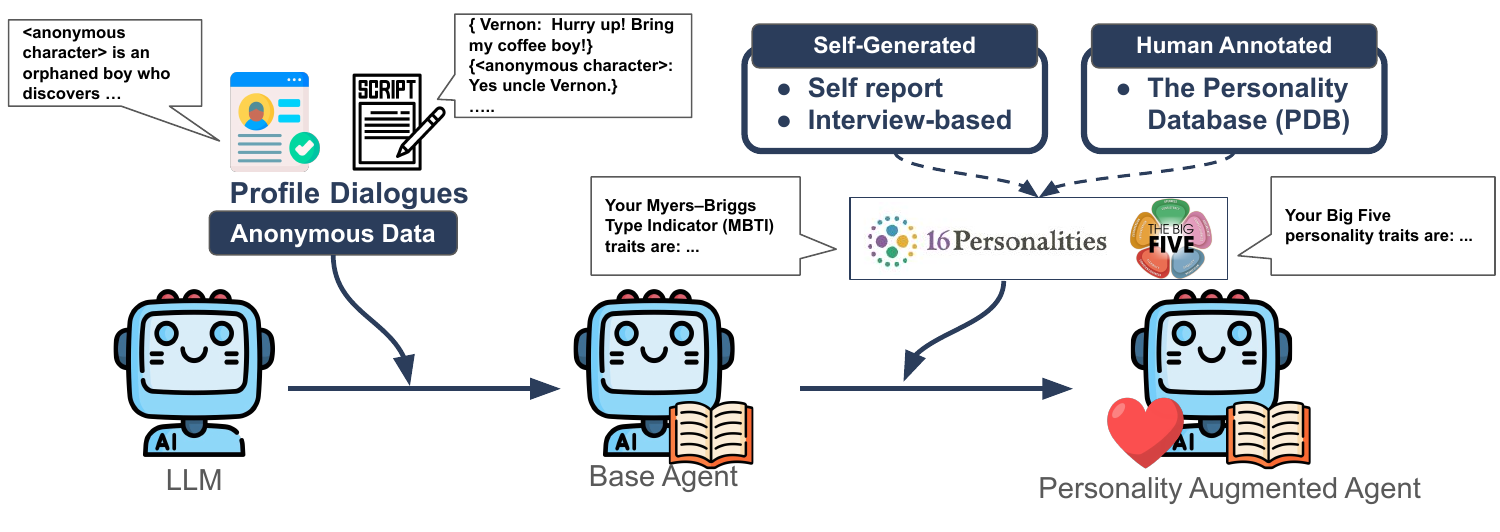}
\caption{The framework of constructing a personality augmented role-playing agent under anonymous scenarios.}
\label{figure:framework}
\end{figure*}

In addition, several studies have explored the simulation of real individuals, including tasks such as survey completion \citep{park2024generative}, social media post generation \citep{huang2024orca}, and message response imitation \citep{shi2025impersona}. However, evaluation methodologies for LLMs impersonating lesser-known real humans remain limited, largely due to the difficulty of collecting high-quality, large-scale data about lesser-known real individuals. Moreover, well-known fictional characters are often well-represented in LLM pretraining corpora, granting models substantial prior knowledge about their personas. Consequently, high proficiency in role-playing these famous figures does not necessarily generalize to the simulation of lesser-known individuals, as models are unfamiliar with the these people.

To address this imbalance in prior knowledge, we propose an anonymized role-playing evaluation method in which character identities are concealed. Under this setting, LLMs must rely solely on the provided character descriptions rather than leveraging pretraining-based prior knowledge. This design mitigates confounding effects stemming from memorized character information and enables a more rigorous assessment of role-playing performance. Importantly, it also facilitates a more principled extrapolation from benchmark results to real-world human impersonation scenarios. An overview of our research framework is illustrated in \autoref{figure:framework}.

While anonymization ensures rigorous evaluation, it also challenges the LLM's ability to maintain character fidelity without prior knowledge. To address this gap, we investigate whether incorporating personality information can bolster role-playing performance under anonymized conditions. Since personality serves as a structured descriptor of human behavioral tendencies \citep{mccrae1997personality}, it provides a precise mechanism for guiding role-play in the absence of character names. Drawing on prior literature \citep{wang2024incharacter}, we compare multiple approaches to acquiring character personality information, including human-annotated data from The Personality Database (PDB)\footnote{\url{https://www.personality-database.com/}}
 and model-generated personalities.

Our experiments demonstrate that incorporating personality information consistently improves role-playing performance. Notably, personality self-generated by the model achieve performance comparable to those derived from human annotations. These gains are robust across different model architectures and multiple benchmark datasets.

Importantly, whereas much of the prior work reports results solely on self-constructed datasets \citep{zhang2025roleplot, gao2025tailorrpa, liu2024roleagent}, our study emphasizes cross-dataset validation. By evaluating across diverse benchmarks and experimental conditions, we show that anonymization significantly reduces model performance in fictional character role-playing, while personality augmentation consistently improves role fidelity. This cross-benchmark consistency strengthens the empirical validity of our findings.
The contributions of this paper are as follows:
\begin{compactitem}
    \item We propose a new evaluation setting for role-play: anonymizing character names in the dataset so that the model must rely solely on the provided role descriptions in the prompt. Experimental results show that anonymization degrades previously reported role-play performance, suggesting that character names carry significant implicit information for LLMs. Therefore, anonymized evaluation offers a fairer and more generalizable assessment.
    \item We investigate whether personality augmentation enhances LLM-based role-playing by enabling the model to generate personality traits from character information. Across multiple benchmark datasets, we show that incorporating personality guidance consistently improves role-playing performance.
    \item  We further compare personalities generated by the model with human-annotated data sourced from PDB. Experimental results across multiple benchmarks indicate that self-generated personalities achieve performance comparable to human-annotated ones, suggesting that effective personality guidance can be obtained without relying on external annotation resources.
\end{compactitem}

\section{Anonymous Role-Playing Evaluation}
%\subsection{Overview of Role-Playing Agents}

RPAs leverage LLMs to simulate a character's knowledge~\citep{lu2024large}, speaking style~\citep{wang2024rolellm}, and behavior~\citep{park2023generative}.
These systems can be categorized by their {\bf simulation target} and {\bf simulation scale}  \citep{chen2025designguidelinerpaevaluation}. Simulation Target refers to whether the RPA simulate an individual \citep{xu2024mindecho, liu2024roleagent,wang2024rolellm} or a group \citep{suzgun2024meta, park2023generative}. Simulation Scale refers to the number of targets simulated in a given context. This study focuses on RPAs with an individual target and a single agent to simulate character persona, where the target is an existing character from literature, TV shows, or animation.

Two main approaches are used to construct character personas: (1) \textbf{Supervised fine-tuning (SFT)}, where models are trained on character-specific Q\&A pairs \citep{wang2024rolellm, shao2023characterllmtrainableagentroleplaying, wang2025coser, yu2024dialogueprofiledialoguealignmentframework}, even enabling smaller models to capture characters' style. (2) \textbf{In-context learning (ICL)}, which conditions models with prompts containing character profiles \citep{tu2024charactereval} or memory from source texts \citep{li2023chatharuhi,liu2024roleagent, gao2025tailorrpa}. Many recent studies adopt ICL for its flexibility in defining roles and inputs, whereas SFT-based RPAs are limited to specific training data \citep{wang2024rolellm} and generalize poorly across roles or benchmarks. We therefore choose ICL, which allows personality information to be directly incorporated into RPAs.

In the first phase of this study, we assess whether LLMs memorize characters through name exposure. We evaluate the effect of anonymizing character names and show that removing names reduces performance. Moreover, under this setting, evaluation results obtained from benchmarks that use fictional characters as role-playing targets can be more reliably generalized to other impersonation scenarios. These include cases in which the target persona is not represented in the model’s pretraining data, such as lesser-known real humans or situations where the role-playing target is a specific social group.

\begin{table}[t!]
    \centering
    \caption{Datasets for role-playing evaluation. ``Char.'' denotes Charactereval, while ``RoleAg.'' denotes RoleAgentBench. Numbers in parentheses indicate the number of characters available on PDB.}
    \label{tab:dataset_statistic}
    \fontsize{10pt}{10pt}\selectfont
    \setlength{\tabcolsep}{5pt} % Reduce column spacing
    \begin{tabular}{llrr}
    \toprule
    \bf Dataset & \bf Task & \bf \#Ques. & \bf \#Char. (PDB)\\
    \midrule
    Char. (CH) & Resp. Gen. & 8,032 & 77 (43)\\
    \midrule
    RoleAg. & General  & 1,005  & 54 (39)\\
    (EN \& CH) & Summary & 462 & 54 (39)\\
    \bottomrule
    \end{tabular}
\end{table}

\begin{table*}[t!]
    \centering
    \caption{CharacterEval scores in Character Consistency dimension of three models: original vs. anonymous settings. $^\dagger$ denotes a statistically significant deterioration ($p < 0.05$).}
    \fontsize{9pt}{9.5pt}\selectfont
    \setlength{\tabcolsep}{2pt} % Reduce column spacing
    \label{tab:vs_original}
    \begin{tabular}{lccccc|c}
    \toprule
%    \bf CharacterEval
    & \bf Know-Exposure	& \bf Know-Accuracy & \bf Know-Hallucination & \bf Persona-Behavior & \bf Persona-Utterance & \bf Avg.\\
        \midrule
\multicolumn{5}{l}{\texttt{gemini-2.0-flash}}	\\	
Original & 2.667~~~~~ & 3.187~~~~ & 3.286~~~~ & 3.787~~~~ & 3.323~~~~ & 3.251~~~~~\\							
Anonymous & 2.656 (-)~~ & 3.178 (-) & 3.250 (-) & 3.783 (-) & 3.277 (-) & 3.229 (-)\\
    \midrule
\multicolumn{5}{l}{\texttt{llama-3.1-405B-instruct}}	\\	
Original & 2.037~~~~~ & 3.013~~~~~ & 3.030~~~~~ & 3.266~~~~ & 3.190~~~~ & 2.907~~~~~\\		
Anonymous & 2.097~~~~~ & 3.029~~~~~ & 3.046~~~~~ & 3.205 (-) & 3.187 (-) & 2.913~~~~~\\
    \midrule
\multicolumn{5}{l}{\texttt{gpt-4o}}	\\	
Original & 2.509~~~~ & 3.029~~~~ & 3.021~~~~ & 2.839~~~~ & 3.024~~~~~ & 2.884~~~~~\\							
Anonymous & 2.461 (-) & 2.986 (-) & 2.969 (-) & 2.607$^\dagger$ (-) & 2.938$^\dagger$ (-) & 2.792$^\dagger$ (-)\\
\bottomrule
    \end{tabular}
\end{table*}

\begin{figure*}[t!]
\centering
    \includegraphics[width=0.70\textwidth]{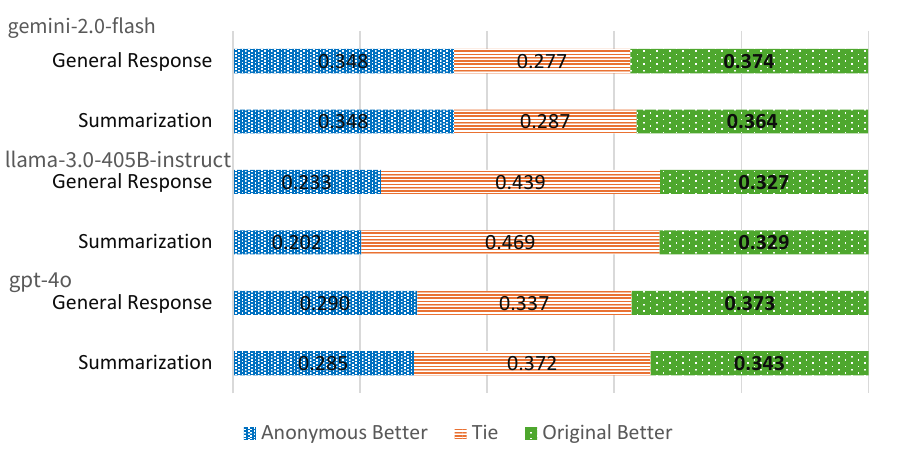}
    \caption{Paired response evaluation on RoleAgentBench: win rates of original vs. anonymous settings.}
    \label{fig:pair-anonymous}
\end{figure*}

\subsection{Anonymizing Process}
To create anonymous scenarios during experiments, we replaced all simulated character names in the prompt with the token \texttt{<anonymous character>}. For example, when simulating Harry Potter, if the profile reads ``\textit{Harry Potter's friends include Ron and Hermione.},'' it becomes ``\texttt{<anonymous character>'s} \textit{friends include Ron and Hermione.}''. Since \textit{Ron} and \textit{Hermione} were not the simulation target of the RPA, they remained un-anonymized. The LLM was thus required to role-play without access to the simulated character’s name, relying solely on information in the prompt. After the RPA generated its responses, we restored the original character names to avoid any impact on evaluation results.

\subsection{Datasets and Evaluation Methods}
We evaluate our approach using two RPA benchmarks that cover diverse question formats, evaluation methods, and languages: \textbf{CharacterEval} \citep{tu2024charactereval} and \textbf{RoleAgentBench} \citep{liu2024roleagent}. The statistic of two benchmarks is shown in \autoref{tab:dataset_statistic}.

\paragraph{CharacterEval} is a Chinese role-playing dialogue dataset designed to evaluate model-generated responses conditioned on a character’s profile and dialogue history, covering 3 dimensions and 11 objectives: \textbf{Conversational Ability} (Fluency, Coherency, Consistency), \textbf{Role-Playing Attractiveness} (Human-Likeness, Communication Skills, Expression Diversity, Empathy), and \textbf{Character Consistency} (Knowledge-Exposure, Knowledge-Accuracy, Knowledge Hallucination, Persona-Behavior, Persona-Utterance). The responses are scored on a continuous scale from 1 to 5 using a reward model trained in the original study.

 \paragraph{RoleAgentBench} includes roles from both Chinese and English scripts and provides each character’s profile along with dialogue history involving other characters. It consists of \textbf{General Response} and \textbf{Summarization} two tasks, which measure general communication skills and memory system effectiveness respectively\footnote{Since \textbf{Self-Knowledge} and \textbf{Reaction}, the other two tasks in RoleAgentBench, are less relevant to the role-playing capabilities we focus on, we exclude these two tasks from our experiments.}.
Following the original paper, we adopt the LLM-as-a-judge evaluation, where the LLM uses the character profile and the dataset’s gold answer to select the output that best matches the character's traits and computes win rates \citep{dubois2025lengthcontrolledalpacaevalsimpleway}. We focus on the publicly available subset of the benchmark.

\subsection{Models}
\label{sec:models}
Previous studies have shown that older models like GPT-3.5 perform significantly worse than GPT-4 in role-playing tasks \citep{suzgun2024meta}. Since this study focuses on personality effects and personality is more abstract than factual information, LLMs must possess sufficient role-playing capability to reflect personality input impact. Therefore, we conduct experiments using recent API-based models: \texttt{gemini-2.0-flash} and \texttt{gpt-4o}. We also include the open-source model \texttt{llama-3.1-405B-instruct} to compare performance between open-source and closed-source models.

For RoleAgentBench evaluation, we use \texttt{gpt-4o-mini} as the evaluator model. The evaluation is conducted in a pairwise manner, where the prompt order is swapped and run twice. A response is considered a win only if it wins both runs; one win and one loss results in a tie, and two losses count as a loss. The pairwise evaluation prompt template, adapted from \citep{liu2024roleagent}, is shown in \autoref{fig:pairwise_prompt}.

\subsection{Anonymous Evaluation Results}
\autoref{tab:vs_original} presents experimental results comparing the original and anonymous versions of the CharacterEval dataset. We focus on the \textit{Character Consistency} dimension, as it best reflects the model's ability to portray specific fictional characters with high fidelity. Overall, responses in the original setting receive higher scores than those in the anonymous setting, suggesting that the reward model finds them more aligned with the intended character traits. Comprehensive experimental results are detailed in \autoref{tab:vs_original_full} of Appendix, which reveals that the Original setting consistently outperforms the Anonymous setting across other objectives as well.

Similar trends are observed in RoleAgentBench, as shown in \autoref{fig:pair-anonymous}. Under LLM-based pairwise evaluation, the original responses exhibit consistently higher win rates than the anonymized responses, indicating better alignment with the ground-truth role-playing behavior.

This performance gap suggests that results reported in previous work may have benefited from models’ memorization of character names. Employing an anonymized evaluation setting enables a fairer comparison of models’ ability to perform role-playing based solely on the information provided in the prompt, thereby making the evaluation results more generalizable to scenarios beyond fictional character role-playing. Therefore, we conduct all subsequent experiments under this setting.

\section{Personality-Augmented RPAs}

\subsection{Personality of Role-Playing Agents}
In psychology perspective, personality are often assessed through standardized questionnaires. Users indicate their level of agreement with each item, and a final score is computed \citep{john1991big, pittenger1993utility}. According to \citet{wang2024incharacter}, there are three main approaches for obtaining the personality of RPAs:
(1) \textbf{Self-report}, where the RPA is prompted with the scale items and possible responses, selecting an option directly to produce a score;
(2) \textbf{Interview-based}, which reformulates scale items into open-ended questions, allowing the RPA to respond freely, after which an evaluator LLM assigns personality scores based on the responses;
(3) \textbf{Crowdsourced data}, which retrieves personality from the PDB website based on crowd-sourced votes for the target character.

\begin{table*}[t!]
    \centering
    \caption{CharacterEval with MBTI or Big Five integration across three dimensions: \emph{Conversational Ability, Role-playing Attractiveness}, and \emph{Character Consistency}. $^\dagger$ denotes a significant improvement over the original condition ($p < 0.05$); \textbf{Bold} fonts indicate the best performance and \underline{underline} fonts indicate the second best one.}
    \label{tab:personality}
    \setlength{\tabcolsep}{5pt}
    \begin{tabular}{clccc|ccc|ccc}
    \toprule
   & & \multicolumn{3}{c}{\texttt{gemini-2.0-flash}} & \multicolumn{3}{c}{\texttt{llama-3.1-405B-inst.}} & \multicolumn{3}{c}{\texttt{gpt-4o}}  \\
   &   & \bf Conv.	& \bf Attr. & \bf Char. & \bf Conv.	& \bf Attr. & \bf Char. & \bf Conv.	& \bf Attr. & \bf Char.\\
      \midrule
\multirow{4}{*}{MBTI} & RPA original &	3.871 &	3.468 &	3.254 &	3.829 &	3.079 & \bf 2.898 &	\underline{3.542} & 2.969 &	2.801\\
& + SelfReport	& \bf 3.924$^\dagger$ & 3.489 & \underline{3.282} & \underline{3.846} & 3.082 & \underline{2.887} & 3.538 & \bf 3.176$^\dagger$ & \bf 2.975$^\dagger$\\
& + Interview & 3.919\ & \underline{3.499} & 3.262 &\bf 3.858 & \bf 3.104	& 2.884 & 3.527 & \underline{3.168}$^\dagger$ & \underline{2.960}$^\dagger$\\
& + PDB & \underline{3.920} & \bf 3.512 & \bf 3.289 & \bf 3.858 & \underline{3.102} & \underline{2.887}	& \bf 3.556$^\dagger$ & 3.052$^\dagger$ & 2.875$^\dagger$\\
\midrule
\multirow{4}{*}{Big 5} & RPA original & 3.951 &  \underline{3.549} & 3.372 & \bf 3.972 & 3.131 & \bf 3.064 & 3.718 & 3.017 & 2.873 \\
& + SelfReport & \underline{3.997} &  3.509 & \bf 3.427 & \underline{3.971} & \bf 3.203 & \bf 3.064 & 3.701 & \bf 3.086 & \underline{2.955}\\
& + Interview & 3.992 & 3.491 & 3.376 & 3.970 & 3.175 & 3.012 & \underline{3.723} & \underline{3.082} & \bf 2.972\\
& + PDB & \bf 4.057 & \bf 3.557 & \underline{3.408} & 3.961 & \underline{3.197} & \underline{3.046} & \bf 3.736 & 3.056 & 2.932\\
\bottomrule
    \end{tabular}
\end{table*}

\begin{figure*}[t!]
\centering
    \includegraphics[width=\textwidth]{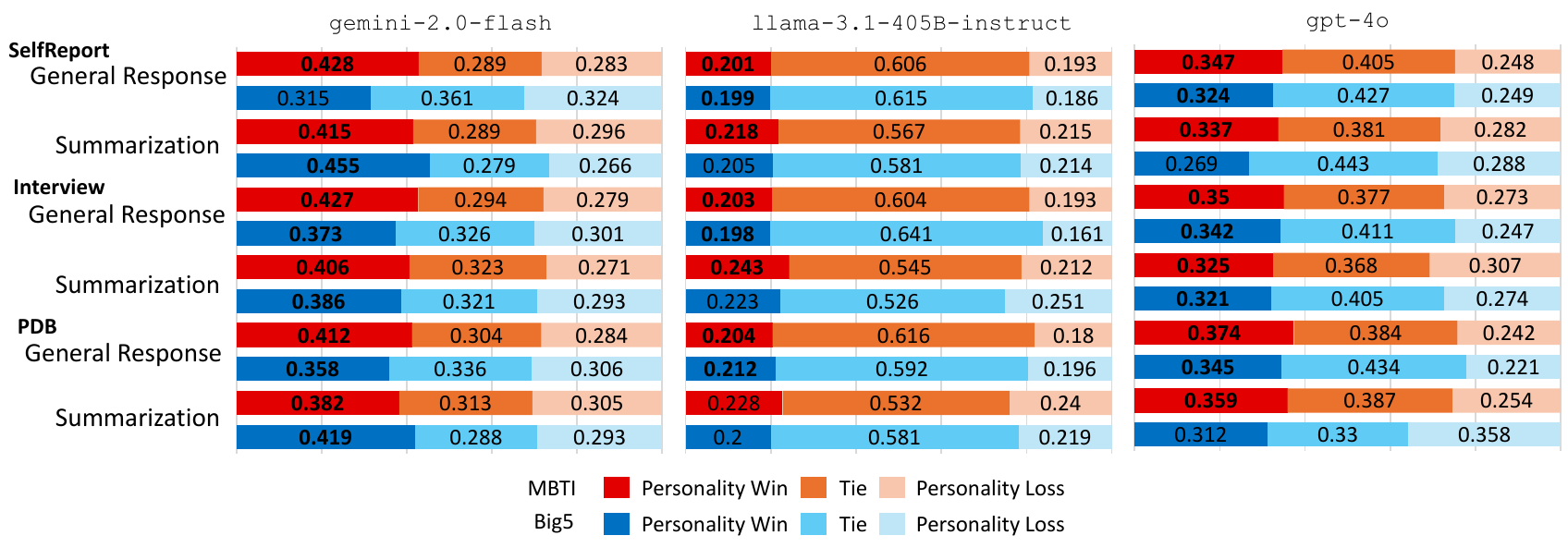}
    \vspace{-4mm}
    \caption{Pairwise comparison on RoleAgentBench: personality integration vs. original condition.}
    \label{fig:winrate-mbti}
\end{figure*}

Prior research has explored methods for inferring personality in RPAs \citep{wang2024incharacter, serapio2023personality}. However, relatively little work has examined whether incorporating personality information can improve role-playing performance. A few recent studies \citep{tu2023characterchat, huang2024orca} touch on personality-enhanced role-playing, but their objectives differ from ours. Specifically, \citet{tu2023characterchat} focused on increasing the diversity of LLM-generated characters through personality information, whereas \citet{huang2024orca} inferred user personality from social media posts, which has relatively low alignment with personality results directly derived from standardized assessment. Moreover, leveraging the personality information of fictional characters to improve role-playing fidelity has not yet been systematically studied.

In this work, we address this gap by investigating whether personality information derived from these various acquisition methods consistently improves role-playing performance. We focus on evaluating the comparability between model-generated personalities (Self-report and Interview-based) and human-annotated data (PDB). By incorporating each source into RPAs, we systematically analyze whether self-generated personality information can serve as a reliable alternative to human annotations across multiple models and benchmarks.

\subsection{Personality Enhancement}
In this study, we use the 16Personalities test\footnote{\url{https://www.16personalities.com/}} to assess the personality of RPAs. This scale is based on the Myers-Briggs Type Indicator (MBTI) \citep{pittenger1993utility} and is widely adopted in both academic and public settings. It is a type-based test consisting of 60 items and yields a result composed of four dimensions. The output is a four-letter personality type (e.g., INTJ). 

Following the implementation \citep{tu2024charactereval}, we obtain each character’s personality type using both the self-report and interview-based methods. For the interview-based approach, we employ \texttt{gemini-2.0-flash} as the evaluator to infer personality types. Additionally, we retrieve personality types from the PDB when available. We then provide the four-letter personality type and the definitions of each category as additional input to the RPA during role-playing. The prompt design is structured according to prior work \citep{wang2024rolellm}, as detailed in \autoref{fig:prompt_template} of Appendix.

\subsection{Personality-Augmented RPA Results}
To fairly compare the three personality sources, our evaluation focuses on characters available in the PDB, and we report the mean score over three runs to ensure reliability. 
The results show that, across both benchmarks, augmenting the prompt with personality information consistently improves performance across most dimensions for all models. This enhancement is evidenced by higher absolute scores (\autoref{tab:personality}) and win rates that significantly exceed loss rates (\autoref{fig:winrate-mbti}), indicating that personality-augmented RPAs generate behaviors more aligned with their target characters. The comprehensive experimental results for each objective of CharacterEval are presented in \autoref{tab:personality_full_MBTI}.

Notably, all three personality sources enhance RPA performance to a similar degree. This finding suggests that self-generated personality traits are a practical and effective alternative to PDB data, as they offer broader applicability to any character while maintaining strong performance. 

Our experiments further demonstrate that even when the model is unaware of the character’s identity, self-generated personality information can significantly improve role-playing performance. These improvements are consistently observed across multiple models and benchmark datasets, highlighting the flexibility of our approach and its ability to robustly enhance RPA performance under diverse settings. Importantly, this suggests that our method is applicable to scenarios in which the model lacks prior familiarity with the target persona, such as impersonating real individuals who are not represented in its pretraining data.

\section{Discussions and Analysis}
\subsection{Generalization to Big Five Personality Inventory}
\label{sec:ablation}
To further verify the effectiveness of personality information in enhancing RPA performance, we adopt the \textbf{Big Five Inventory (BFI)} \citep{john1991big}, a widely used personality assessment instrument. The Big Five personality are measured as continuous scores across five dimensions, reflecting individual personality characteristics and allowing for more fine-grained distinctions. In contrast, while the MBTI framework is less sensitive to subtle variations, it is generally easier to interpret. Therefore, we evaluate both frameworks to show that widely used personality models, regardless of their structural differences, consistently improve RPA performance.

In our experiments, we provide RPAs with the BFI scores for each character to construct personality-augmented agents. To ensure a fair comparison, we evaluate only characters with available PDB personality annotations. However, Big Five data is sparse, covering only 9 characters in CharacterEval and 25 in RoleAgentBench. This limited sample size accounts for the differing ``Original'' baseline results across the two benchmarks.

The results, presented in \autoref{tab:personality} and \autoref{fig:winrate-mbti}, demonstrate that BFI also yields performance gains, suggesting that the benefits of incorporating personality information are robust across different frameworks. However, MBTI outperforms BFI in terms of enhancement magnitude. We hypothesize that this disparity stems from the prevalence of MBTI in public online discourse; for instance, the PDB contains more MBTI-related character analyses than BFI. Consequently, LLMs may have developed a superior ability to map MBTI traits to specific character behaviors during. The underlying mechanisms behind this performance disparity warrant further investigation. The comprehensive experimental results for each objective of CharacterEval are presented in \autoref{tab:personality_full_BFI} of Appendix.
\subsection{Greater Improvement from Stronger Personality Traits}

To further investigate the influence of personality on RPAs, we identified characters with distinctive personality profiles to evaluate whether such traits correlate with greater performance improvements. We define a distinctive personality profile as having MBTI scores either above $60\%$ or below $40\%$ across all four dimensions; only characters meeting this criterion were included in the analysis. 

We conducted a granular evaluation of \texttt{gemini-2.0-flash}, as it not only achieved the highest average score on CharacterEval but also demonstrated the most significant win-rate margin in the personality-augmented experiments on RoleAgentBench. We hypothesize that \texttt{gemini-2.0-flash} is better equipped to leverage personality information for role-playing; consequently, we expect the presence of a distinctive personality to yield even more substantial performance gains. The distribution of these characters across each condition is detailed in \autoref{tab:num_sig_character}. We specifically focus on the performance gains localized to this character subset, contrasting their improvements with the performance of the full dataset to highlight the efficacy of personality augmentation.

\begin{table}[t!]
    \centering
    \caption{Distribution of characters with distinctive personality types across different experimental conditions.}
    \label{tab:num_sig_character}
    \fontsize{10pt}{10pt}\selectfont
    \setlength{\tabcolsep}{5pt} % Reduce column spacing
    \begin{tabular}{lccc}
    \toprule
    \bf Dataset  & \bf Self-report & \bf Interview & \bf PDB\\
    \midrule
    CharacterEval & 22 & 20 & 37\\
    RoleAgentBench  & 3  & 14  & 33\\
    \bottomrule
    \end{tabular}
\end{table}

\autoref{tab:ablation_significant} reports the performance differences when considering only roles with strong personality traits, where positive values imply that stronger personality information enhances role-playing capabilities more. We observe that characters with distinctive personality types generally yield greater performance gains. These results suggest that when a character exhibits distinct personality tendencies, RPAs can more effectively leverage personality cues to generate behaviors that align more closely with the original character, thereby providing clearer evidence of the importance of personality information in LLM-based role-playing.

\begin{table}[t!]
    \centering
    \caption{Additional performance improvement attributed to distinctive personalities with \texttt{gemini-2.0-flash}. RoleAgentBench values denote win rate minus loss rate.}
    \label{tab:ablation_significant}
    \fontsize{10pt}{10pt}\selectfont
    \setlength{\tabcolsep}{4pt} % Reduce column spacing
    \begin{tabular}{lccc|cc}
    \toprule
    & \multicolumn{3}{c}{\bf CharacterEval} & \multicolumn{2}{|c}{\bf RoleAgentBench}\\
%   \cline{2-4} \cline{5-6}
    & \bf Conv. & \bf Attr. & \bf Char.	& \bf Gen. Resp. & \bf Sum.\\
    \midrule
    SelfR. & -.013 & \bf +.021 & -.012 & \bf +.076 & \bf +.122\\
    Inter. & \bf +.138 & -.088 & \bf +.067 &\bf +.055 &\bf +.020\\
    PDB & \bf +.152 & \bf +.291 & \bf +.162 & \bf +.101 & -.149\\
    \bottomrule
    \end{tabular}
\end{table}

\begin{table*}[t!]
    \centering
    \caption{Sample responses with different methods.}
    \label{tab:qualitative}
    \small
    \begin{tabular}{p{15.5cm}}
    \toprule
    {\bf Question (from McGonagall):}
    Mr. Malfoy, do you recall your behavior during our first flying lesson and how you provoked Mr. Potter, which led to him being discovered as a potential Quidditch player for Gryffindor?\\
    \midrule
    {\bf Original RPA (Malfoy):} \colorbox{yellow}{(Scoffs)} Of course I remember. Longbottom was being an idiot, as usual. It was hilarious. \colorbox{yellow}{And Potter? He should be thanking me!} I inadvertently gave him a chance to show off. Without me, he'd just be another nobody. Besides, it's not my fault Potter can't resist a challenge. He practically begged me to goad him. \\
    \midrule
     {\bf Anonymized RPA:}
     Yes, I recall Longbottom making a fool of himself. And I certainly remember showing Longbottom's Remembrall to Potter, wondering if the fat lump could even reach it on the roof. What's it to you? Potter being a Seeker hardly seems like *my* fault. He's still a mudblood-loving Gryffindor, isn't he?\\
     \midrule
     {\bf + SelfReport Personality (INTJ):} 
     \colorbox{yellow}{(Scoffs, crossing his arms)} Recall it? Of course, I recall it. Potter showing off, thinking he's so special. \colorbox{yellow}{Longbottom was being an idiot, dropping that stupid Remembrall.} I was just having a bit of fun. Potter deserved to be knocked down a peg or two. He got lucky, that's all. And if he hadn't been so busy trying to be a hero, maybe the stupid lump wouldn't have made such a fool of himself.\\
    \bottomrule
    \end{tabular}
\end{table*}
\subsection{Qualitative Analysis}
In addition to benchmark scores, we conducted a qualitative analysis of RPAs' responses across different settings. Sample responses are shown in \autoref{tab:qualitative}. Prior to incorporating personality information, responses in the Original setting showed more character-specific tone and emotion compared to the Anonymized setting, likely due to the LLM leveraging prior knowledge when aware of the character's name. After adding personality information, the responses better reflected the character's tone, emotional state, and included actions, resulting in more vivid and expressive outputs. This suggests that even without access to the character’s name, personality information can provide meaningful cues that help the RPA generate more lifelike and engaging role-play behavior.

\begin{figure*}[t!]
\centering
    \includegraphics[width=0.8\textwidth]{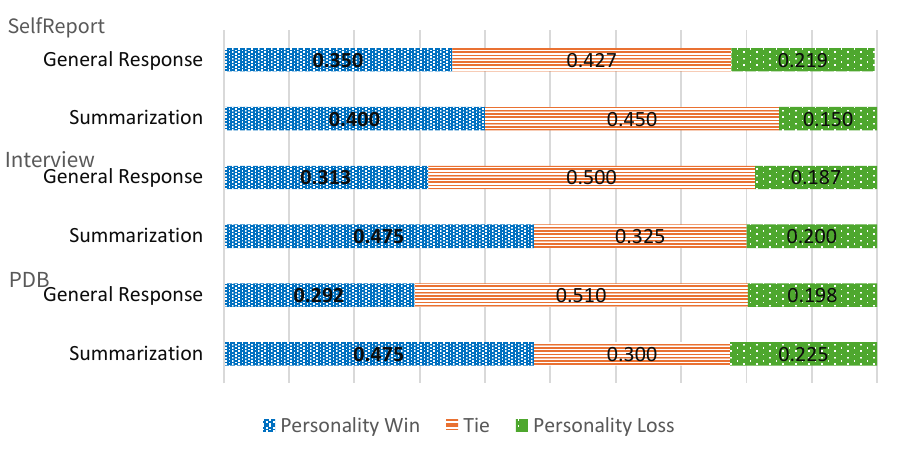}
    \vspace{-5mm}
    \caption{Human pairwise comparison against the original condition.}
    \label{fig:human_eval}
\end{figure*}

\subsection{Human Evaluation}
To further validate these findings, we conducted a human evaluation on RoleAgentBench. We used the generation results of \texttt{gemini-2.0-flash} and focused on Harry Potter data to ensure evaluator familiarity. 
Two psychology graduate students with expertise in the Harry Potter series performed pairwise comparisons. A response was deemed a ``win'' if both evaluators preferred it, a ``loss'' if neither did, and a ``tie'' if they disagreed. 

The evaluation covering role-playing for 5 characters from the Harry Potter series. Each evaluator assessed 120 and 288 pairwise comparisons for the summary and general response tasks, respectively. In terms of inter-rater agreement, Cohen’s kappa was 0.415 for the summary task and 0.308 for the general response task, corresponding to fair to moderate agreement according \citep{mchugh2012interrater}.

The results, presented in \autoref{fig:human_eval}, confirm that personality-augmented RPAs outperform the original condition under human judgment, aligning with the findings from our LLM-based evaluations.

\section{Conclusion}
This work introduces an anonymous role-play setting and proposes the incorporation of personality information into RPAs to enhance performance. We compare multiple personality sources to assess their impact. Our experiments are conducted across multiple benchmarks, languages, and model architectures, yielding robust and consistent results.

Our findings show that the anonymous setting generally reduces RPA performance, highlighting LLMs’ sensitivity to character names. This suggests that conventional role-play evaluation may partially rely on prior knowledge of characters. By contrast, anonymous evaluation provides a fairer assessment framework and produces results that are more generalizable to scenarios beyond fictional character role-playing.

We further observe that incorporating personality information consistently improves role-playing performance. Notably, self-generated personality information achieves performance comparable to human annotations, demonstrating a flexible and scalable approach that can be applied across diverse settings.

\section*{Limitations}
This study has several limitations that open avenues for future research. First, although our anonymization strategy effectively reduces name cues, models may still infer character identities from other descriptive signals, such as the names of interlocutors, nicknames mentioned in dialogue, or distinctive and widely recognized character traits. Future work could develop more comprehensive anonymization techniques to further mitigate potential confounding effects.

Second, although most models in our study exhibit performance gains from personality augmentation, the magnitude of improvement varies across model architectures. Further experiments on a broader range of models are needed to better understand what factors contribute to these differences. Identifying the mechanisms underlying such variability remains an important direction for future research.

Finally, while our experiments confirm that both personality frameworks enhance RPA performance, it remains unclear why the magnitude of improvement varies across frameworks, or how these disparate information sources can be optimally combined to achieve peak performance. Addressing these questions represents an important direction for future research.

\section*{Acknowledgements}
This work was financially supported by the National Science and Technology Council (NSTC) in Taiwan, under Grant 112-2223-E-002-012-MY5.
We thank the National Center for High performance Computing of National Institutes of
Applied Research (NIAR) in Taiwan for providing
computational and storage resources.

\bibliographystyle{acl_natbib}
\bibliography{custom}

\appendix

\section{Prompts for Constructing RPA and Pairwise Evaluation}
\label{sec:appendix_prompt}
\autoref{fig:prompt_template} depicts the prompt configuration for the personality-augmented RPA. The underlying architecture adapts to varying benchmark instructions, with comprehensive details provided in the codebase. Within this framework, personality information is operationalized using four-letter MBTI types and five-dimensional BFI scores. For the non-augmented condition, the prompts are simplified by truncating all content prior to the string: "Your MBTI traits are: {personality info}." Furthermore, the prompt of LLM-based pairwise evaluation protocol for RoleAgentBench is detailed in \autoref{fig:pairwise_prompt}.

\begin{figure*}[t!]
\footnotesize
\begin{tcolorbox}[
  colback=gray!10,     % Background color
  colframe=gray!80,    % Border color
  fonttitle=\bfseries, % Bold font for title
  coltitle=white,      % Title color
  rounded corners,     % Rounded corners
  width=\linewidth,    % Full width
  boxrule=0.5mm        % Thickness of the border
]
\texttt{System prompt:} \\
\{character's profile\}\\
\\
\texttt{User prompt:} \\
Myers–Briggs Type Indicator (MBTI) is a self-report questionnaire. The test assigns a binary value to each of four categories: introversion or extraversion, sensing or intuition, thinking or feeling, and judging or perceiving. Extraverts (also often spelled extroverts) are outward-turning and tend to be action-oriented, enjoy more frequent social interaction, and feel energized after spending time with other people. Introverts are inward-turning and tend to be thought-oriented, enjoy deep and meaningful social interactions, and feel recharged after spending time alone.. People who prefer sensing tend to pay a great deal of attention to reality. Those who prefer intuition pay more attention to things like patterns and impressions. People who prefer thinking place a greater emphasis on facts and objective data. Those who prefer feeling are more likely to consider people and emotions when arriving at a conclusion. Those who lean toward judging prefer structure and firm decisions. People who lean toward perceiving are more open, flexible, and adaptable. \\
Your MBTI traits are:
\{personality info\}. \\
Please anwser the following questions based on character's information.\\
The following is the dialogue record of the role:\\
\{dialogue history between the character and other agents\} \\
\{task instruction and question\}
\end{tcolorbox}
\vspace{-1mm}
\caption{Prompt template for personality-augmented RPA.}
\label{fig:prompt_template}
\vspace{-3mm}
\end{figure*}
\begin{figure*}[t!]
\footnotesize
\begin{tcolorbox}[
  colback=gray!10,     % Background color
  colframe=gray!80,    % Border color
  fonttitle=\bfseries, % Bold font for title
  coltitle=white,      % Title color
  rounded corners,     % Rounded corners
  width=\linewidth,    % Full width
  boxrule=0.5mm        % Thickness of the border
]
\texttt{System prompt:} \\
You are a role-playing performance comparison assistant. You should rank the conditions based on the role characteristics and text quality of their responses. The rankings are then output using Python dictionaries and lists. \\
\\
\texttt{User prompt:} \\
The conditions below are to play the role of \{character name\}. The role description of \{character name\} is \{character’s profile\}. I need to rank the following conditions based on the two criteria and the reference answer below: \\
1. Which one has more pronounced role speaking style, and speaks more in line with the role description. The more distinctive the speaking style, the better. \\
2. Which one's output contains more knowledge and memories related to the role; the richer, the better. (If the question contains reference answers, then the role-specific knowledge and memories are based on the reference answer.) \\
The question provided to each condition is: \{benchmark question\} \\
The reference answer of the question is: \{benchmark answer\} \\
The respective answers from the conditions to this question are: \\
\text{[\{"condition": "condition 1", "answer": \{answer 1\}\},
\{"condition": "condition 2", "answer": \{answer 2\}\}]} \\
Now, based on the above two criteria and the reference answer, please rank the conditions. Avoid any positional biases and ensure that the order in which the responses are presented does not influence your decision. Do not favor certain model names.
Then, use a list containing the condition's name, its rank, and the reason for its ranking to return the results, i.e., please ensure to use the following format to return the results: \\
\text{[\{"condition": <condition-name>, "reason": <rank-reason>, "rank": <condition-rank>\},} \\
\text{\{"condition": <condition-name>, "reason": <rank-reason>, "rank": <condition-rank>\}]} \\
Your answer must be a valid Python list of dictionaries to ensure I can directly parse it using Python. Do not include any extraneous content! Please provide a ranking that is as accurate as possible and aligns with the intuition of most people.
\end{tcolorbox}
\vspace{-1mm}
\caption{Prompt template for RoleAgentBench pair-wise evaluation.}
\label{fig:pairwise_prompt}
\vspace{-3mm}
\end{figure*}

\section{Full results of experiments}
\label{sec:appendix_full}
\autoref{tab:vs_original_full}, \autoref{tab:personality_full_MBTI}, and \autoref{tab:personality_full_BFI} report the dimension-wise performance on CharacterEval, providing extended versions of the results summarized in \autoref{tab:vs_original} and \autoref{tab:personality}. As illustrated in \autoref{tab:vs_original_full}, the anonymous setting generally leads to a performance degradation across most objectives. This suggests that the absence of explicit character names negatively impacts the model's ability to portray well-known fictional characters effectively.

Furthermore, \autoref{tab:personality_full_MBTI} and \autoref{tab:personality_full_BFI} demonstrate that incorporating personality information enhances role-playing performance across the majority of objectives, regardless of the personality source. Notably, the most significant improvements are observed in \textbf{Human-likeness, Expression Diversity, and Persona-Behavior}. These results indicate that personality profiles enable RPAs to generate more diverse, human-like, and behaviorally consistent responses.
\begin{table*}[h]
\centering
\caption{Full comparison between Original and Anonymous settings on CharacterEval. \underline{Underlined} values indicate that the anonymous setting resulted in a lower score compared to the original, and $^\dagger$ denotes a statistically significant deterioration ($p < 0.05$). The colors \textcolor{cyan}{Cyan}, \textcolor{teal}{Teal}, and \textcolor{purple}{Purple} represent the Conversational Ability, Role-Playing Attractiveness, and Character Consistency dimensions, respectively.}
\resizebox{\textwidth}{!}{
\begin{tabular}{lcc|cc|cc}
    \toprule
    & \multicolumn{2}{c}{gemini-2.0-flash} & \multicolumn{2}{c}{llama-3.1-405B-instruct} & \multicolumn{2}{c}{gpt-4o} \\
    \midrule
        & \textbf{Original} & \textbf{Anonymous} & \textbf{Original} & \textbf{Anonymous} & \textbf{Original} & \textbf{Anonymous} \\
    \midrule
         \textcolor{cyan}{Fluency} & 3.779 & \underline{3.744} & 3.713 & \underline{3.682} & 3.516 & \underline{3.455}  \\
         \textcolor{cyan}{Coherency} & 4.089 & 4.099 & 4.046 & \underline{4.032} & 3.874 & \underline{3.831} \\
         \textcolor{cyan}{Consistency} & 3.914 & \underline{3.883} & 3.928 & \underline{3.893} & 3.495 & \underline{3.406}$^\dagger$ \\
         \textcolor{teal}{Human-likeness} & 3.679 & \underline{3.637} & 3.743 & \underline{3.693} & 3.155 & \underline{3.099} \\
         \textcolor{teal}{Communication Skills} & 3.618 & \underline{3.588} & 3.049 & 3.057 & 3.571 & \underline{3.466}$^\dagger$ \\
         \textcolor{teal}{Expression Diversity} & 3.255 & \underline{3.245} & 2.623 & \underline{2.574} & 2.236 & \underline{2.103}$^\dagger$ \\
         \textcolor{teal}{Empathy} & 3.410 & \underline{3.399} & 3.142 & \underline{3.106} & 3.306 & \underline{3.274} \\
         \textcolor{purple}{Know-Exposure} & 2.667 & \underline{2.656} & 2.037 & 2.097 & 2.509 & \underline{2.461} \\
         \textcolor{purple}{Know-Accuracy} & 3.187 & \underline{3.178} & 3.013 & 3.029 & 3.029 & \underline{2.986} \\
         \textcolor{purple}{Know-Hallucination} & 3.286 & \underline{3.250} & 3.030 & 3.046 & 3.021 & \underline{2.969} \\
         \textcolor{purple}{Persona-Behavior} & 3.787 & \underline{3.783} & 3.266 & \underline{3.205} & 2.839 & \underline{2.607}$^\dagger$ \\
         \textcolor{purple}{Persona-Utterance} & 3.323 & \underline{3.277} & 3.190 & \underline{3.187} & 3.024 & \underline{2.938}$^\dagger$ \\
    \bottomrule
    \end{tabular}}
    \label{tab:vs_original_full}
\end{table*}
\vspace{-0.5em}
\begin{table*}[h]
\centering
\caption{Full comparison between w/ MBTI Personality and w/o Personality in anonymous scenario on CharacterEval. “Original” refers to prompts without personality information. Bold values indicate that the score in the w/ personality setting is higher than in the original, and $^\dagger$ denotes a statistically significant improvement ($p < 0.05$). The colors \textcolor{cyan}{Cyan}, \textcolor{teal}{Teal}, and \textcolor{purple}{Purple} represent the Conversational Ability, Role-Playing Attractiveness, and Character Consistency dimensions, respectively.} 
\resizebox{\textwidth}{!}{
\begin{tabular}{lcccc|cccc|cccc}
    \toprule
    & \multicolumn{4}{c}{gemini-2.0-flash} & \multicolumn{4}{c}{llama-3.1-405B-instruct} & \multicolumn{4}{c}{gpt-4o} \\
    \midrule
        & \textbf{Original} & \textbf{SelfReport} & \textbf{Interview} & \textbf{PDB} & \textbf{Original} & \textbf{SelfReport} & \textbf{Interview} & \textbf{PDB} & \textbf{Original} & \textbf{SelfReport} & \textbf{Interview} & \textbf{PDB} \\
    \midrule
         \textcolor{cyan}{Fluency} & 3.676 & \textbf{3.749} & \textbf{3.754} & \textbf{3.749} & 3.598 & \textbf{3.631} & \textbf{3.641} & \textbf{3.642} & 3.418 & 3.329 & 3.351 & 3.411\\
         \textcolor{cyan}{Coherency} & 4.084 & \textbf{4.120} & \textbf{4.116} & \textbf{4.109} & 4.022 & 4.018 & \textbf{4.040} & \textbf{4.031} & 3.803 & 3.754 & 3.730 & 3.793\\
         \textcolor{cyan}{Consistency} & 3.852 & \textbf{3.903} & \textbf{3.888} & \textbf{3.903} & 3.865 & \textbf{3.890} & \textbf{3.892} & \textbf{3.901} & 3.405 & \textbf{3.532}$^\dagger$& \textbf{3.502} & \textbf{3.464}\\
         \textcolor{teal}{Human-likeness} & 3.632 & \textbf{3.643} & \textbf{3.666} & \textbf{3.683} & 3.692 & \textbf{3.710} & \textbf{3.758} & \textbf{3.781} & 3.094 & \textbf{3.271}$^\dagger$& \textbf{3.261}$^\dagger$& \textbf{3.150}$^\dagger$\\
         \textcolor{teal}{Communication Skills} & 3.618 & \textbf{3.660} & \textbf{3.650} & \textbf{3.682} & 3.044 & \textbf{3.054} & \textbf{3.073} & \textbf{3.045} & 3.475 & 3.404 & 3.399 & 3.458\\
         \textcolor{teal}{Expression Diversity} & 3.213 & \textbf{3.222} & \textbf{3.260} & \textbf{3.248} & 2.465 & 2.443 & 2.456 & 2.457 & 2.020 & \textbf{2.938}$^\dagger$& \textbf{2.919}$^\dagger$& \textbf{2.365}$^\dagger$\\
         \textcolor{teal}{Empathy} & 3.411 & \textbf{3.430} & \textbf{3.419} & \textbf{3.453} & 3.116 & \textbf{3.123} & \textbf{3.129} & \textbf{3.126} & 3.286 & 3.092& 3.093 & 3.235\\
         \textcolor{purple}{Know-Exposure} & 2.759 & \textbf{2.786} & 2.748 & \textbf{2.771} & 2.124 & 2.046 & 2.064 & 2.057 & 2.493 & \textbf{2.564} & \textbf{2.568} & \textbf{2.517}\\
         \textcolor{purple}{Know-Accuracy} & 3.189 & \textbf{3.195} & \textbf{3.199} & \textbf{3.203} & 3.023 & \textbf{3.029} & \textbf{3.026} & 3.008 & 3.021 & 2.920 & 2.913 & 2.965\\
         \textcolor{purple}{Know-Hallucination} & 3.285 & \textbf{3.331} & \textbf{3.306} & \textbf{3.363}$^\dagger$& 3.051 & \textbf{3.090} & 3.038 & \textbf{3.071} & 3.010 & \textbf{3.028} & 2.995 & \textbf{3.024}\\
         \textcolor{purple}{Persona-Behavior} & 3.730 & \textbf{3.777} & \textbf{3.733} & \textbf{3.773} & 3.117 & 3.091 & 3.100 & 3.106 & 2.541 & \textbf{3.337}$^\dagger$& \textbf{3.313}$^\dagger$& \textbf{2.869}$^\dagger$\\
         \textcolor{purple}{Persona-Utterance} & 3.309 & \textbf{3.323} & \textbf{3.325} & \textbf{3.336} & 3.175 & \textbf{3.177} & \textbf{3.194} & \textbf{3.196} & 2.938 & \textbf{3.023} & \textbf{3.010} & \textbf{2.998} \\
    \bottomrule
    \end{tabular}
    }
    \label{tab:personality_full_MBTI}
\end{table*}
\begin{table*}[h]
\centering
\renewcommand{\arraystretch}{1.4}
\caption{Full Comparison between w/ BFI Personality and w/o Personality in anonymous scenario on CharacterEval. “Original” refers to prompts without personality information. Bold values indicate that the score in the w/ personality setting is higher than in the original, and $^\dagger$ denotes a statistically significant improvement over the original condition ($p < 0.05$). The colors \textcolor{cyan}{Cyan}, \textcolor{teal}{Teal}, and \textcolor{purple}{Purple} represent the Conversational Ability, Role-Playing Attractiveness, and Character Consistency dimensions, respectively.}
\resizebox{\textwidth}{!}{
\begin{tabular}{lcccc|cccc|cccc}
    \toprule
    & \multicolumn{4}{c}{gemini-2.0-flash} & \multicolumn{4}{c}{llama-3.1-405B-instruct} & \multicolumn{4}{c}{gpt-4o} \\
    \midrule
        & \textbf{Original} & \textbf{SelfReport} & \textbf{Interview} & \textbf{PDB} & \textbf{Original} & \textbf{SelfReport} & \textbf{Interview} & \textbf{PDB} & \textbf{Original} & \textbf{SelfReport} & \textbf{Interview} & \textbf{PDB} \\
    \midrule
         \textcolor{cyan}{Fluency} & 3.556 & \textbf{3.706} & \textbf{3.641} & \textbf{3.798} & 3.687 & 3.606 & 3.670 & 3.556 & 3.400 & 3.378 & \textbf{3.469} & 3.350\\
         \textcolor{cyan}{Coherency} & 4.230 & \textbf{4.261} & \textbf{4.285} & \textbf{4.253} & 4.157 & \textbf{4.216} & \textbf{4.163} & \textbf{4.205} & 4.055 & 4.045 & \textbf{4.058} & \textbf{4.077}\\
         \textcolor{cyan}{Consistency} & 4.067 & 4.024 & 4.050 & \textbf{4.122} & 4.072 & \textbf{4.090} & \textbf{4.076} & \textbf{4.122} & 3.699 & 3.681 & 3.643 & \textbf{3.781}\\
         \textcolor{teal}{Human-likeness} & 3.807 & 3.696 & 3.697 & 3.608 & 3.787 & 3.763 & 3.757 & 3.763 & 3.151 & \textbf{3.167} & \textbf{3.250} & \textbf{3.195}\\
         \textcolor{teal}{Communication Skills} & 3.811 & 3.738 & 3.733 & \textbf{3.863} & 3.265 & \textbf{3.317} & 3.232 & \textbf{3.268} & 3.553 & \textbf{3.644} & 3.528 & \textbf{3.591}\\
         \textcolor{teal}{Expression Diversity} & 2.926 & \textbf{2.991} & \textbf{2.988} & \textbf{3.090} & 2.196 & \textbf{2.303} & \textbf{2.297} & \textbf{2.338} & 1.797 & \textbf{1.951} & \textbf{1.989} & \textbf{1.852}\\
         \textcolor{teal}{Empathy} & 3.652 & 3.613 & 3.548 & \textbf{3.665} & 3.357 & \textbf{3.428} & \textbf{3.415} & \textbf{3.420} & 3.569 & \textbf{3.580} & 3.560 & \textbf{3.587}\\
         \textcolor{purple}{Know-Exposure} & 3.200 & \textbf{3.273} & 3.105 & 3.184 & 2.607 & 2.566 & 2.529 & 2.559 & 2.940 & \textbf{2.995} & \textbf{2.978} & 2.921\\
         \textcolor{purple}{Know-Accuracy} & 3.320 & \textbf{3.352} & \textbf{3.365} & \textbf{3.399} & 3.267 & \textbf{3.291} & 3.195 & \textbf{3.297} & 3.100 & \textbf{3.132} & \textbf{3.144} & 3.099\\
         \textcolor{purple}{Know-Hallucination} & 3.364 & \textbf{3.423} & \textbf{3.420} & \textbf{3.420} & 3.297 & 3.222 & 3.191 & \textbf{3.230} & 2.841 & \textbf{2.993} & \textbf{3.035} & \textbf{3.037}\\
         \textcolor{purple}{Persona-Behavior} & 3.545 & \textbf{3.649} & \textbf{3.563} & \textbf{3.616} & 2.828 & \textbf{2.906} & \textbf{2.875} & 2.826 & 2.329 & \textbf{2.541} & \textbf{2.649} & \textbf{2.521}\\
         \textcolor{purple}{Persona-Utterance} & 3.433 & \textbf{3.437} & 3.425 & 3.420 & 3.321 & \textbf{3.333} & 3.269 & 3.315 & 3.154 & 3.115 & 3.055 & 3.087 \\
    \bottomrule
    \end{tabular}
    }
    \label{tab:personality_full_BFI}
\end{table*}
\end{document}